\documentclass[11pt]{article}
\usepackage{epsfig}
\usepackage{amssymb,amsmath}
\usepackage{epigraph}
\usepackage{xcolor}
\usepackage[lined,boxed,commentsnumbered]{algorithm2e}
\usepackage{graphicx}
\usepackage{multirow}
\usepackage{siunitx,booktabs}
\usepackage{newtxtext,newtxmath}
\usepackage{mdframed}

\usepackage{booktabs}
\usepackage{siunitx}

\begin{document}

\begin{center}
{\LARGE
CAISSON: Concept-Augmented Inference Suite of Self-Organizing Neural Networks
}

\vskip1.0cm
{\Large Igor Halperin\footnote{Fidelity Investments. E-mail: igor.halperin@fmr.com. Opinions expressed here are author's own, and do not necessarily represent views of his employer. A standard disclaimer applies.}} \\
\vskip0.5cm
\today \\

\vskip1.0cm
{\Large Abstract:\\}
\end{center}

\vskip0.3cm

\parbox[t]{\textwidth}{
We present CAISSON, a novel hierarchical approach to Retrieval-Augmented Generation (RAG) that transforms traditional single-vector search into a multi-view clustering framework. At its core, CAISSON leverages dual Self-Organizing Maps (SOMs) to create complementary organizational views of the document space, where each view captures different aspects of document relationships through specialized embeddings. The first view processes combined text and metadata embeddings, while the second operates on metadata enriched with concept embeddings, enabling a comprehensive multi-view analysis that captures both fine-grained semantic relationships and high-level conceptual patterns. This dual-view approach enables more nuanced document discovery by combining evidence from different organizational perspectives. To evaluate CAISSON, we develop SynFAQA, a framework for generating synthetic financial analyst notes and question-answer pairs that systematically tests different aspects of information retrieval capabilities. Drawing on HotPotQA's methodology for constructing multi-step reasoning questions, SynFAQA generates controlled test cases where each question is paired with the set of notes containing its ground-truth answer, progressing from simple single-entity queries to complex multi-hop retrieval tasks involving multiple entities and concepts. Our experimental results demonstrate substantial improvements over both basic and enhanced RAG implementations, particularly for complex multi-entity queries, while maintaining practical response times suitable for interactive applications.
}

\begin{center}
\vskip0.2cm
\includegraphics[height=6.5cm]{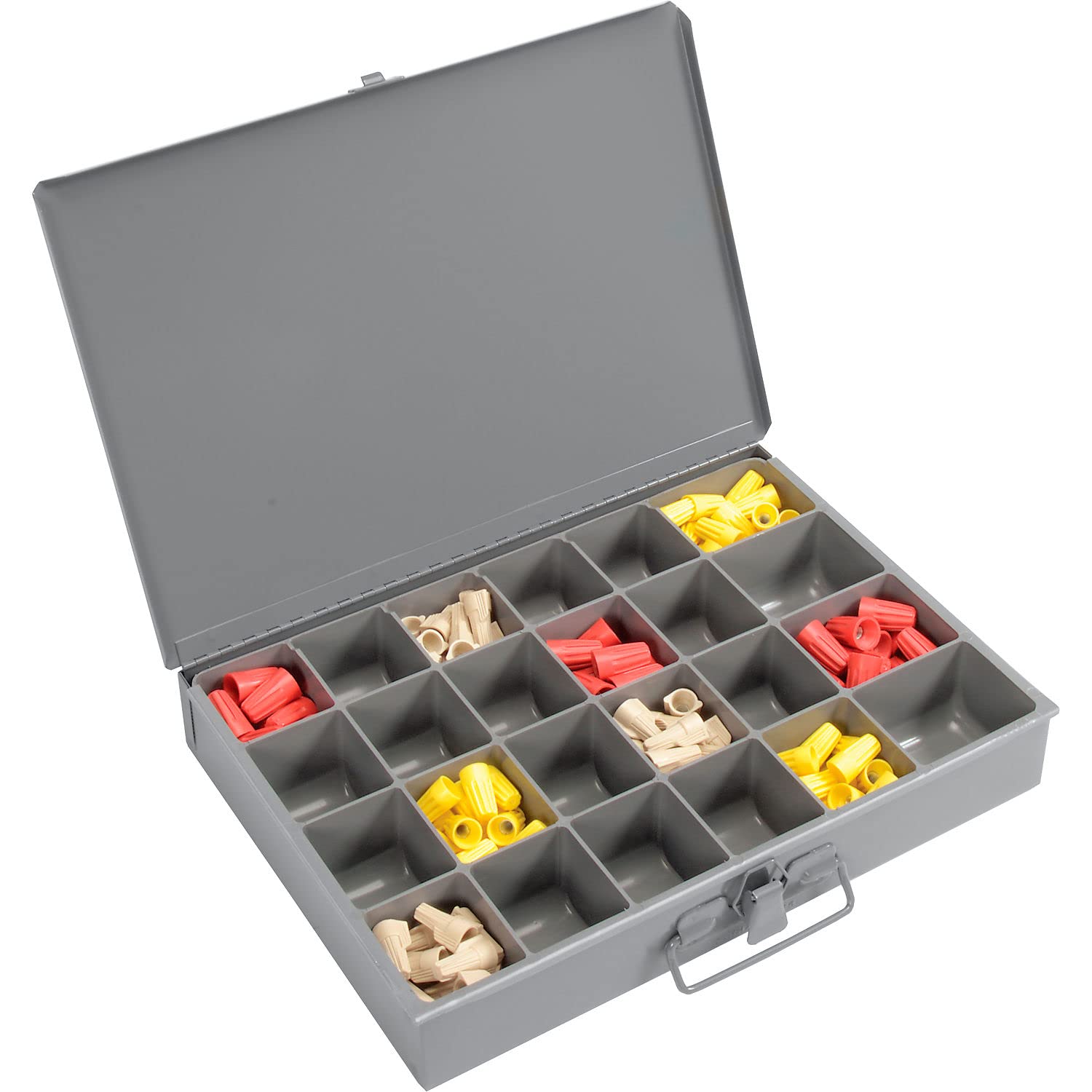}
\end{center}

\newcounter{helpfootnote}
\setcounter{helpfootnote}{\thefootnote} 
\renewcommand{\thefootnote}{\fnsymbol{footnote}}
\setcounter{footnote}{0}
\footnotetext{I would like to thank Claude Sonnet 3.5 for code development, suggestions on bug fixes and methodology improvements, as well as for assistance with writing the manuscript.
Special thanks are to Vijay Saraswat for constructive criticism of the first draft or this paper.}     

\renewcommand{\thefootnote}{\arabic{footnote}}
\setcounter{footnote}{\thehelpfootnote} 

\newpage

\section{Introduction}
\label{sect_Intro}

Retrieval-Augmented Generation (RAG) has emerged as a common industry-wide method to enhance the capabilities of large language models (LLMs) by allowing them to access and incorporate external knowledge sources during text generation ~\cite{Lewis2020rag}. Despite extensive research and development in this area, current RAG systems relying on straightforward vector similarity searches applied to chunks of text often face several challenges for complex queries, including potential loss of relevant information, handling of multi-entity queries, and incorporation of domain-specific concepts and relationships. While various improvements have been proposed in the literature, most solutions maintain the fundamental single-vector similarity paradigm, limiting their ability to capture multiple aspects of document relationships simultaneously.

This paper addresses these limitations by introducing CAISSON (Concept-Augmented Inference Suite of Self-Organizing Neural Networks), a novel approach that reimagines RAG architecture through the lens of multi-view clustering. CAISSON combines modern transformer-based embeddings with classical Self-Organizing Maps (SOMs)~\cite{Kohonen2001self} to create complementary organizational views of the document space. Each view captures different aspects of document relationships: one processes combined text and metadata embeddings for semantic organization, while the other operates on metadata enriched with concept embeddings for thematic clustering. This multi-view approach creates a rich organizational framework where documents are simultaneously clustered according to different similarity criteria, enabling more sophisticated retrieval than traditional single-vector approaches. Our experimental results demonstrate the effectiveness of this approach, achieving up to 148\% improvement in retrieval accuracy over baseline RAG 
implementations, and maintaining strong performance even for complex four-entity queries.

CAISSON implements this multi-view vision through two specialized SOMs that process different aspects of document representations. Each SOM creates a topologically-organized network of retrieval nodes, effectively clustering similar documents according to different similarity criteria. The complementary views enable both semantic matching through text-metadata embeddings and conceptual matching through concept-metadata embeddings. During retrieval, evidence from both views is combined to provide a more comprehensive assessment of document relevance. This architecture can be viewed as a 'swarm of specialized RAGs' working in parallel across different organizational views of the document space.

To rigorously evaluate the effectiveness of this multi-view architecture, particularly its ability to handle queries of varying complexity, we need a systematic evaluation framework that can test both simple retrieval capabilities and complex multi-entity retrieval tasks. Traditional benchmarks, typically focused on single-aspect retrieval or general question-answering tasks, are inadequate for assessing the unique capabilities of multi-view document organization. To address this need, we developed SynFAQA (Synthetic Financial Analyst Questions and Answers generator), a comprehensive evaluation framework for financial document retrieval systems. Drawing inspiration from HotPotQA's methodology for constructing multi-step reasoning questions~\cite{Yang2018hotpotqa} and incorporating complexity scaling principles from DeepMind's Michelangelo framework~\cite{DeepMind2024michelangelo}, SynFAQA first generates a collection of synthetic analyst notes that reflect real-world characteristics such as industry sector distributions, market capitalization weightings, and typical patterns in multi-entity coverage. These notes then serve as the foundation for generating controlled test cases of increasing complexity, where each question is paired with the set of notes containing its ground-truth answer. The resulting framework enables systematic evaluation of capabilities ranging from simple fact retrieval from a single document to complex multi-hop inference requiring information synthesis across multiple documents. While initially developed to evaluate CAISSON, SynFAQA represents a broader contribution to the field of financial natural language processing.

Our comprehensive evaluation demonstrates several key strengths of the multi-view clustering approach:
\begin{itemize}
    \item Superior retrieval performance across query complexities, with Mean Reciprocal Rank (MRR) of 0.5231 compared to 0.2106 for basic RAG
    \item Exceptional handling of multi-entity queries, achieving peak performance (MRR 0.5703) for two-ticker queries
    \item Robust scaling to complex queries, maintaining strong performance (MRR 0.4366) even with four tickers
    \item Efficient query processing with sub-second response times suitable for interactive applications
\end{itemize}

Our paper is organized as follows. After summarizing our key contributions and reviewing related work in the rest of Section~\ref{sect_Intro}, in Section~\ref{sect_CAISSON} we present the CAISSON model, detailing our novel dual-SOM architecture, the embedding approaches for different views, and the mechanisms for combining evidence across views. Section~\ref{sect_CAISSON_test} presents our experimental framework, including the construction of synthetic analyst notes and training methodology. In Section~\ref{sect_NoteRetriever}, we introduce NoteRetriever, our fast document search algorithm implemented in CAISSON, detailing the query processing pipeline and scoring mechanisms that make efficient retrieval possible. Section~\ref{sect_SynFAQA} presents SynFAQA, our comprehensive Q\&A generation and evaluation framework for testing the performance of language models for financial analysis. The SynFAQA dataset is then used in Section~\ref{sect_CAISSON_evaluation} to evaluate CAISSON's performance relative to both basic and enhanced RAG implementations. We conclude in Section~\ref{sect_Summary} with a summary of our findings and discussion of potential extensions to other domains with rich entity relationships and concept hierarchies.

\subsection{Key Contributions}

This paper makes several contributions to the field of document retrieval and RAG systems:

\begin{itemize}
\item \textbf{Novel RAG Architecture:} CAISSON introduces a fundamental reimagining of RAG systems through a self-organizing network of specialized retrieval nodes where:
    \begin{itemize}
        \item Each node functions as a specialized RAG instance for specific semantic or conceptual patterns
        \item The global topology preserves relationships between different specializations
        \item The dual paths enable separate optimization of semantic and conceptual retrieval
    \end{itemize}

\item \textbf{Hybrid Classical-Modern Approach:} CAISSON demonstrates how classical neural architectures can be effectively combined with modern transformer-based embeddings to create a robust retrieval system that:
    \begin{itemize}
        \item Leverages the topological preservation properties of SOMs
        \item Utilizes the semantic understanding capabilities of modern transformers
        \item Creates a scalable, hierarchical document organization
    \end{itemize}

\item \textbf{Multi-View Document Representation:} We develop a comprehensive embedding framework that captures different aspects of document relationships through:
    \begin{itemize}
        \item Semantic path combining text and entity information
        \item Concept path integrating domain knowledge and metadata
        \item Specialized handling of multi-entity and multi-concept documents
    \end{itemize}

\item \textbf{Scalable Implementation and Evaluation Framework:} We provide:
    \begin{itemize}
        \item Computationally efficient implementation scaling linearly with dataset size
        \item SynFAQA: a novel evaluation framework for financial document retrieval
        \item Comprehensive benchmarking against baseline RAG approaches
    \end{itemize}
\end{itemize}

\subsection{Related Work}

\subsubsection{Multi-View Learning and Document Organization}

Recent advances in multi-view clustering have demonstrated the power of leveraging complementary information across different representations of the same objects. As reviewed by Chen et al.~\cite{Chen2022multiview}, multi-view clustering approaches can effectively combine different aspects of data to improve clustering quality, particularly when different views provide complementary information. While much of the recent work has focused on deep learning-based approaches, we demonstrate that classical clustering algorithms like Self-Organizing Maps, which have a long history of successful applications in document organization since the 1990s, can be effectively combined with modern transformer-based embeddings to create powerful multi-view retrieval systems. This hybrid approach of combining well-understood classical techniques with state-of-the-art embedding methods offers a promising direction for extending traditional RAG architectures.

%

\subsubsection{Traditional RAG Systems and Their Limitations}

RAG systems combine retrieval-based and generation-based models through a two-stage architecture: (1) an ingestion stage where documents are chunked, embedded, and stored in a vector database, and (2) an inference stage where relevant chunks are retrieved based on query similarity to augment LLM prompts.

Despite widespread adoption, traditional RAG approaches face several fundamental limitations~\cite{Gao2023survey}:
\begin{itemize}
    \item \textbf{Single-View Limitation:} Reliance on single-vector representations fails to capture multiple aspects of document relationships simultaneously
    
    \item \textbf{Information and Context Loss:} Document chunking, while necessary for efficiency, often breaks semantic connections and loses crucial contextual information
    
    \item \textbf{Multi-Entity Challenges:} Standard methods struggle to balance relevance across different entities or concepts in complex queries
\end{itemize}

While various improvements have been proposed, including hierarchical representations and multi-stage pipelines, most solutions maintain the fundamental single-vector similarity paradigm.

\subsubsection{Context-Aware and Concept-Based Approaches}

Recent work has approached RAG limitations from different angles. Anthropic's "Contextual Retrieval"~\cite{Anthropic2024contextual} addresses the context loss problem through context-aware embeddings, prepending chunk-specific context before embedding to maintain broader document relationships. While this effectively preserves context, it operates within a single-view framework.

Complementarily, Kimhi et al.~\cite{Kimhi2022interpreting} proposed transforming embedding spaces into interpretable conceptual spaces through their Conceptualizing Embedding Spaces (CES) methodology. Their approach maps embeddings to human-understandable concepts using a hierarchical ontology derived from structured knowledge sources like Wikipedia's category graph. CES demonstrates that embedding spaces can be effectively organized around semantic concepts, enabling more intuitive navigation and analysis of document relationships. While CES focuses on interpretability of single embeddings, it provides valuable insights into concept-based document representation.

CAISSON synthesizes and extends these insights through a multi-view clustering framework that simultaneously organizes documents along both semantic and conceptual dimensions. This approach enables each view to specialize while maintaining efficient evidence combination during retrieval, offering a promising direction for next-generation RAG systems.

\section{The CAISSON model}
\label{sect_CAISSON}

\subsection{Architectural Overview}

\begin{figure}[t]
    \centering
    \includegraphics[height=8cm,width=0.8\textwidth]{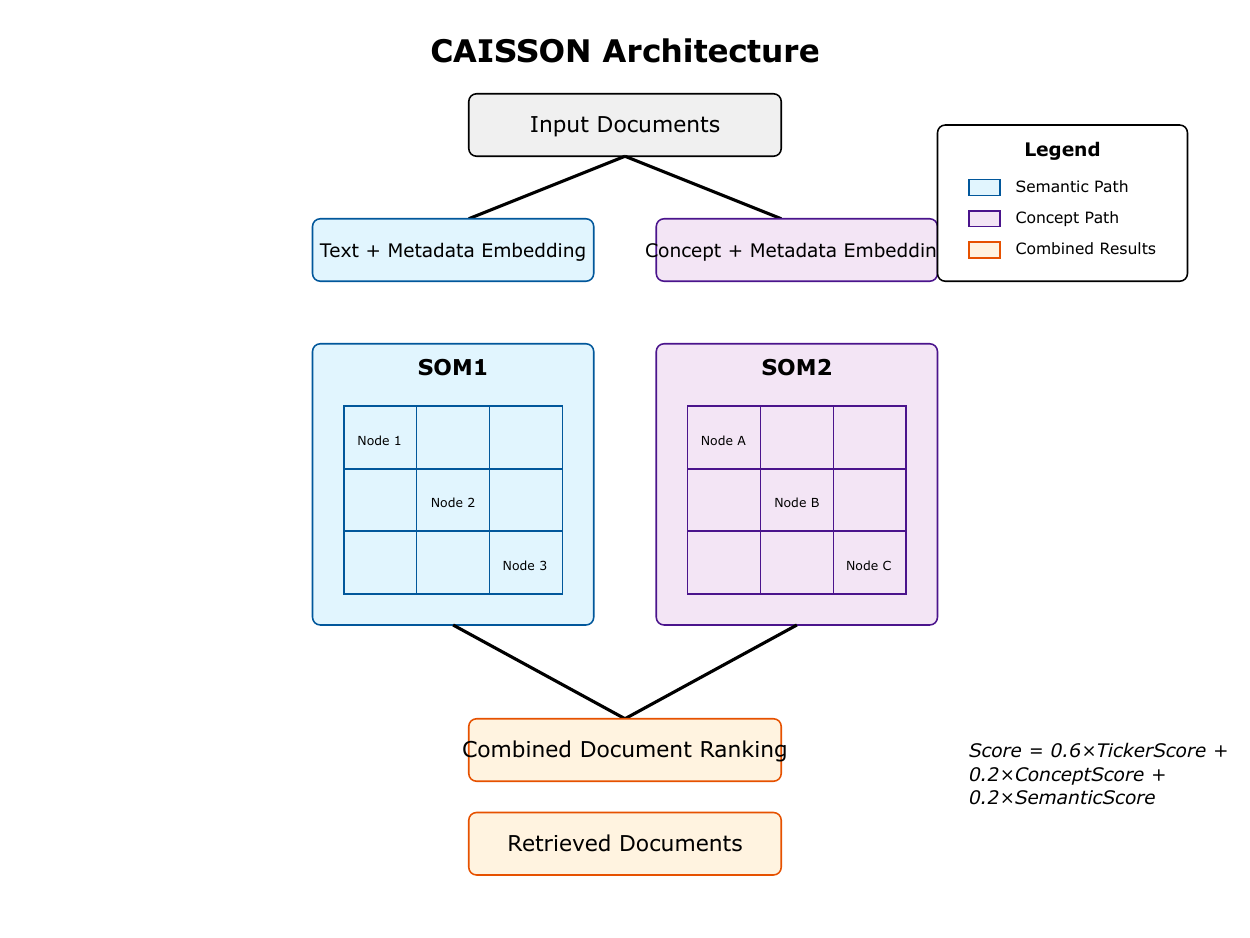}
    \caption{CAISSON's dual-path architecture showing parallel processing through semantic (SOM1) and concept (SOM2) paths. Each SOM node functions as a specialized RAG instance, with the final ranking combining evidence from both paths.}
    \label{fig:architecture}
\end{figure}

CAISSON implements a novel dual-path architecture for document retrieval (Figure~\ref{fig:architecture}). The system processes input documents through two parallel specialized paths, each culminating in a Self-Organizing Map (SOM) that creates a topologically-organized network of retrieval nodes (see in the next subsection for more details on SOMs). Each node within these SOMs effectively functions as a specialized RAG instance, maintaining a local collection of documents that share similar characteristics:

\begin{itemize}
    \item The Semantic Path (SOM1) processes combined text and metadata embeddings, creating nodes that specialize in specific semantic patterns and entity relationships
    \item The Concept Path (SOM2) processes concept and metadata embeddings, creating nodes that specialize in specific thematic and conceptual patterns
\end{itemize}

During retrieval, a query is processed through both paths simultaneously. Each path identifies relevant nodes based on its specialized organization, and documents are retrieved from these nodes' local collections. The final ranking combines evidence from both paths using a weighted scoring function that considers ticker matches, concept similarity, and semantic relevance.

This dual-path architecture offers several advantages over traditional RAG systems:
\begin{enumerate}
    \item Local Specialization: Each node becomes expert at handling specific types of queries
    \item Global Organization: The SOMs maintain topological relationships between different specializations
    \item Complementary Views: The dual paths capture both fine-grained semantic relationships and high-level conceptual patterns
\end{enumerate}


\subsection{Self-Organizing Maps: Implementation Details}

CAISSON extends the classical Self-Organizing Maps (SOMs)~\cite{Kohonen2001self} framework through a novel network architecture that combines traditional SOM learning with modern RAG capabilities. This section details our implementation approach and key architectural innovations.

\subsubsection{Classical SOMs vs. CAISSON's Extended Architecture}

Each SOM in CAISSON consists of an $n \times n$ grid of nodes (with $n=10$ in our implementation), where each node functions as a specialized document retriever that learns to handle specific types of documents. The key innovation in CAISSON lies in how these nodes operate and what information they maintain:

\begin{itemize}
    \item \textbf{Classical SOM Node:}
    \begin{itemize}
        \item Maintains single representative vector $w_i \in \mathbb{R}^d$ (node weight)
        \item Updates $w_i$ during training to represent average characteristics
        \item Discards individual input vectors after updating $w_i$
    \end{itemize}
    
    \item \textbf{CAISSON's Extended Node:}
    \begin{itemize}
        \item Maintains representative vector $w_i \in \mathbb{R}^d$ for training and specialization
        \item Stores complete set of document embeddings $\{E_1, E_2, ..., E_k\} \subset \mathbb{R}^d$ mapped to the node
        \item Functions as specialized vector database for mapped documents
        \item Enables efficient retrieval within node's area of specialization
    \end{itemize}
\end{itemize}

Here $d$ is the embedding dimension. In our implementation, $ d = 434 $ for both SOMs: SOM1 combines a 384-dimensional text and a 50-dimensional entity embeddings, while SOM2 combines a 384-dimensional concept embedding and a 50-dimensional embedding of metadata.\footnote{In our implementation used for numerical experiments in this paper, we proxy document metadata by simply tickers (entities) mentioned in the document, and thus use embeddings of entities (tickers) as embeddings of metadata. In general, the two SOMs of CAISSON can work with vectors of different dimensionalities.}

\subsubsection{Training Process}

The training process adapts the nodes' representative vectors while simultaneously building document collections. We define:
\begin{itemize}
    \item $x(t) \in \mathbb{R}^d$: Input document embedding at time $t$
    \item $w_i(t) \in \mathbb{R}^d$: Representative vector of node $i$ at time $t$
    \item $r_i \in \mathbb{R}^2$: Grid position coordinates of node $i$ in the 2D SOM lattice
    \item $\|\cdot\|$: Euclidean distance norm in the respective space ($\mathbb{R}^d$ or $\mathbb{R}^2$)
    \item $C_i$: Set of document embeddings stored at node $i$
\end{itemize}

For each input document embedding $x(t)$ at time $t$, the process follows these steps:

\begin{enumerate}
    \item \textbf{Competition:} Find Best Matching Unit (BMU) based on minimum distance between input embedding and node representative vectors:
    \begin{equation}
    c = \text{argmin}_i \|x(t) - w_i(t)\|
    \end{equation}
    where $c$ is the index of the winning node.
    
    \item \textbf{Cooperation:} Define neighborhood influence of the winning node $ c $ on all nodes $i $ through spatial proximity in the SOM grid:
    \begin{equation}
    h_{ci}(t) = \alpha(t) \exp\left(-\frac{\|r_c - r_i\|^2}{2\sigma^2(t)}\right)
    \end{equation}
    where:
    \begin{itemize}
        \item $r_c, r_i \in \mathbb{R}^2$: Grid coordinates of winning node $c$ and any node $i$
        \item $\alpha(t) = \alpha_0(1 - \gamma t/T)$: Learning rate decay with hyperparameters $\alpha_0$ and $ \gamma $
        \item $\sigma(t) = \sigma_0\exp(-\lambda t)$: Neighborhood radius decay with hyperparameters $\sigma_0$ and $ \lambda $
        \item $T$: Total number of epochs 
    \end{itemize}
    
    \item \textbf{Adaptation:} Update node representative vectors for all nodes $ i $ according to:
    \begin{equation}
    w_i(t+1) = w_i(t) + h_{ci}(t)[x(t) - w_i(t)]
    \end{equation}
    where the magnitude of update is controlled by the neighborhood function $h_{ci}(t)$.
    
    \item \textbf{Storage:} Add document embedding to BMU's collection:
    \begin{equation}
    C_c = C_c \cup \{x(t)\}
    \end{equation}
\end{enumerate}

Note the distinct roles of vectors and distances in the algorithm:
\begin{itemize}
    \item $w_i(t) \in \mathbb{R}^d$: Representative embeddings in document space
    \item $r_i \in \mathbb{R}^2$: Fixed grid positions in SOM lattice
    \item $\|x(t) - w_i(t)\|$: Similarity in the embedding space
    \item $\|r_c - r_i\|$: Physical distance in SOM grid
\end{itemize}

\subsubsection{Training Monitoring}

The training progress is monitored through quantization error for each SOM:
\begin{equation}
Q_i = \frac{1}{N}\sum_{j=1}^N \|x_j - w_{BMU(x_j)}\|^2
\end{equation}
where $N$ is the number of documents and $BMU(x_j)$ maps document $j$ to its Best Matching Unit index. This error typically shows exponential decay during training, with the majority of organization occurring in the first 50 epochs.

\subsubsection{Retrieval Network Formation}

The final network supports efficient retrieval through a two-stage process:
\begin{enumerate}
    \item Use node representative vectors $\{w_i\}$ to identify relevant specializations
    \item Search document collections $\{C_i\}$ within selected nodes for best matches
\end{enumerate}

CAISSON implements this extended SOM architecture in both paths of its dual architecture, with SOM1 organizing documents based on semantic and entity relationships, while SOM2 focuses on concept and metadata patterns. The specific embedding approaches for each path are detailed in the following subsections.

\subsection{Semantic Path Embeddings (SOM1)}

\subsubsection{Combined Semantic Embedding}
The semantic path combines text and entity embeddings to create a unified document representation. For a document $d$ with text content and a set of entities $\{e_1, \ldots, e_K\}$, the combined embedding is:
\begin{equation}
E_{SOM1}(d) = [E_{text}(d); E_{entities}(\{e_1, \ldots, e_K\})]
\end{equation}
where $E_{text}(d)$ is the text embedding, $E_{entities}$ is the combined entity embedding, and $[\cdot;\cdot]$ denotes vector concatenation. The following subsections detail the construction of each component.

\subsubsection{Text Embedding Model}

CAISSON utilizes the all-MiniLM-L6-v2 Sentence Transformer model for text embeddings, which produces 384-dimensional vectors. For our experimental setting with short analyst notes, this lightweight model provides an effective balance between performance and computational efficiency. The model adequately handles financial terminology and market-specific language without requiring domain-specific fine-tuning.

For applications involving longer documents or more complex text chunks, more sophisticated transformer models might be appropriate, such as MPNet-base (768 dimensions), BERT-large (1024 dimensions), or domain-specific models like FinBERT. However, for our current setting with concise text fragments, our experiments demonstrate that MiniLM's 384-dimensional embeddings capture sufficient semantic information while maintaining computational efficiency.


\subsubsection{Entity (Ticker) Embedding Construction}
The entity embedding framework uses configurable dimensionality, implemented in our experiments with 50-dimensional embeddings. The construction process involves several stages:

\begin{enumerate}
\item Base Initialization: Each ticker receives an initial random embedding:
\begin{equation}
E_{base} \sim \mathcal{N}(0, I_d)
\end{equation}
where $d$ is the embedding dimension (set to 50 in our implementation).

\item Industry Integration: The embedding incorporates industry information through weighted averaging:
\begin{equation}
E_{ticker} = \lambda E_{base} + (1-\lambda) \frac{1}{|S_i|}\sum_{j \in S_i} E_{base}^{(j)}
\end{equation}
where $\lambda = 0.7$ controls the balance between individual and industry-wide characteristics.

\item Market Capitalization Adjustment: A size factor is added using:
\begin{equation}
E_{final} = E_{ticker} + \beta \log(MC) 
\end{equation}
where $MC$ is the market capitalization and $\beta = 0.01$ controls the size effect.

\item Normalization: Final normalization ensures consistent magnitudes:
\begin{equation}
E_{norm} = \frac{E_{final}}{|E_{final}|_2}
\end{equation}
\end{enumerate}

\subsubsection{Multi-Entity Document Representation}
For documents mentioning multiple entities, we implement a simple but effective combination strategy:
\begin{equation}
E_{multi} = \frac{1}{K}\sum_{i=1}^K n_i E_i
\end{equation}
where $K$ is the number of entities, and $ n_i $ is the count of ticker 
$ i $ in the document. This ensures that multi-entity embeddings remain close to their constituents in the embedding space (while tilting towards tickers mentioned more often), an important property for SOM-based clustering.

\subsection{Concept Path Embeddings (SOM2)}

\subsubsection{Combined Concept Embedding}

The concept path combines metadata and concept information into a unified representation. For a document $d$ with metadata $m$ and identified concepts $\{c_1, \ldots, c_N\}$, the combined embedding is:
\begin{equation}
\label{eq_SOM2}
E_{SOM2}(d) = [E_{metadata}(d); E_{concepts}(\{c_1, \ldots, c_N\})]
\end{equation}
where in our implementation we proxy documents' metadata by their entities (tickers) information, i,e. we set 
$ E_{metadata}(d) = E_{entities}(\{e_1, \ldots, e_K\} $ where 
$  E_{entities}(\{e_1, \ldots, e_K\} $ is the same embedding for entity information as constructed above for SOM1.\footnote{We have
also experimented with an alternative specification that adds a one-hot encoding of notes' sentiments, i.e. $   E_{metadata}(d) = \left[ E_{entities}(\{e_1, \ldots, e_K\}; 
E_{sentiment}(d) \right]$, howewer we did not find any useful improvement in performance.}
 The new element in (\ref{eq_SOM2}) is the combined concept embedding $E_{concepts}$. The following subsections detail its construction.

\subsubsection{Concept Embedding Organization}
Our implementation employs a direct approach to concept embedding using the same MiniLM-L6-v2 transformer model used for text embeddings. For each predefined concept in the financial domain (e.g., "Revenue Growth," "Market Share," "Product Launch"), we generate a base embedding:
\begin{equation}
E_c = \text{MiniLM}(c) \quad \forall c \in \mathcal{C}
\end{equation}
where $\mathcal{C}$ is our predefined set of 24 market-relevant concepts. The system maintains a concept similarity cache to optimize repeated computations:
\begin{equation}
s(c_i, c_j) = \cos(E_{c_i}, E_{c_j})
\end{equation}
This base approach could be enhanced using more sophisticated methods. For example, inspired by Anthropic's contextual retrieval methodology~\cite{Anthropic2024contextual}, we could replace our predefined concept set with a more flexible LLM-based approach where concepts emerge from document summaries:
\begin{equation}
E_{concepts} = \text{Embed}(\text{LLM}(\text{"Summarize the key financial concepts in this document: "} + \text{doc}))
\end{equation}
This would allow SOM2 to work with emergent concept representations rather than a fixed concept vocabulary, potentially capturing more nuanced thematic patterns in the documents.

In contrast, Kimhi et al.'s conceptualized semantic space approach~\cite{Kimhi2022interpreting} offers a different direction where documents would be decomposed according to a predefined concept basis. Their method would first establish a fixed set of concept embeddings, then represent documents through their similarities to these basis concepts, effectively projecting documents onto a predefined conceptual space.

These alternatives represent fundamentally different approaches to concept handling: dynamic concept emergence through LLM summarization versus decomposition along fixed concept dimensions. While these approaches offer interesting directions for future work (discussed in Section~\ref{sect_Summary}), in this paper we proceed with our simpler predefined concept approach. This choice enables us to evaluate CAISSON's architectural innovations in isolation, without introducing additional variability from concept identification methods that would complicate the interpretation of results.

\subsubsection{Multi-Concept Document Handling}

CAISSON handles documents with multiple concepts through a two-stage process. First, concepts are identified either through direct matching against our concept dictionary or through inference:
\begin{equation}
C_d = \begin{cases}
\text{metadata.concepts} & \text{if explicitly provided} \\
\text{infer\_concepts}(d) & \text{otherwise}
\end{cases}
\end{equation}

For concept inference, we implement two versions:
\begin{itemize}
\item Version 1: Dictionary-based matching using predefined concept synonyms
\item Version 2: Inference of basis ontological concepts through embedding similarity comparison
\end{itemize}
Our basis implementation that uses synthetic data to illustrate the working of CAISSON relies on the first method for all numerical experiments. This enables filtering out possible errors due to probabilistic inference of concepts, and focusing on the performance of the CAISSON model itself, rather than on the performance of its variable components that can be alternatively used for version 2. 
For real-world applications where the inference of concepts is not amenable to simple NLP-based search methods, version 2 might be a more appropriate option.

The combined concept representation for a document is computed as an unweighted average of individual concept embeddings:
\begin{equation}
E_{concepts} = \frac{1}{|C_d|}\sum_{c \in C_d} E_c
\end{equation}

The concept similarity between documents or between a query and documents is computed using a cached similarity matrix:
\begin{equation}
\text{sim}(C_1, C_2) = \begin{cases}
\frac{|C_1 \cap C_2|}{\max(|C_1|, |C_2|)} & \text{if } C_1 \cap C_2 \neq \emptyset \\
\max\limits_{c_1 \in C_1, c_2 \in C_2} s(c_1, c_2) & \text{otherwise}
\end{cases}
\end{equation}
where $s(c_1, c_2)$ is retrieved from the concept similarity cache when available.

\subsection{Training Process}
CAISSON implements parallel training of both SOMs with independent learning rates but synchronized epochs. For each epoch, documents are processed in batches of size 32. The learning rates decay linearly:
\begin{equation}
\alpha_i(t) = \alpha_i^0 \left(1 - \gamma t/T \right)
\end{equation}
where $i \in {1,2}$ indexes the SOMs, $t$ is the current epoch, $T$ is total epochs, and $ \gamma = 0.8 $ is a hyperparameter controlling the speed of the learning rate decay. Initial rates are set to $\alpha_1^0 = 0.05$ and $\alpha_2^0 = 0.05$.
Training progress is monitored through quantization errors for each SOM:
\begin{equation}
Q_i = \frac{1}{N}\sum_{j=1}^N |x_j - w_{BMU(x_j)}|^2
\end{equation}
where $N$ is the number of documents and $BMU(x_j)$ denotes the Best Matching Unit for document $j$ in SOM $i$.

\subsection{Document Retrieval}
A final ranking process used by trained SOMs to retrieve relevant documents for a given query is described below in Sect.~\ref{sect_NoteRetriever}.

\section{Numerical Experiments with Synthetic Analyst Notes}
\label{sect_CAISSON_test}

\subsection{Synthetic Dataset Construction}
To evaluate CAISSON's performance, we constructed a large synthetic dataset of financial analyst notes that mimics the key characteristics of real market commentary. Our data generation procedure creates structured notes containing text content and associated metadata, with carefully controlled distributions of entities (tickers) and concepts per note.

The dataset construction process incorporates domain knowledge about the S\&P 500 market structure in several ways. First, the ticker universe is organized by industry sectors, with 11 major sectors represented. The ticker selection process is weighted according to approximate market capitalization weights of companies in the S\&P 500, ensuring that larger companies appear more frequently in the dataset. For example, major technology companies like Apple (AAPL) and Microsoft (MSFT) appear in approximately 9-10\% of notes each, while smaller companies typically appear in 2-3\% of notes.

The conceptual coverage of notes is controlled through a predefined set of 24 market-relevant concepts, such as "Earnings beat", "Revenue growth", "Market share gain", and "Technological disruption". To ensure linguistic variety, each concept is associated with multiple synonymous expressions that are randomly selected during text generation. 


\begin{figure}[h]
\centering
\includegraphics[height=6cm, width=13cm]{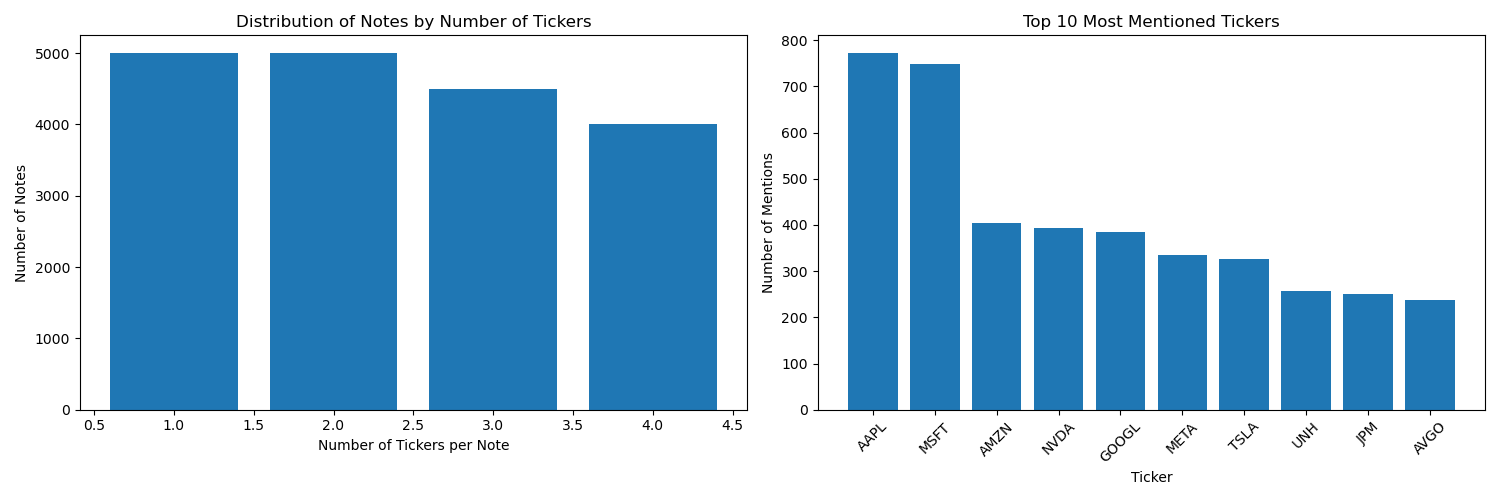}
\caption{Distributions of tickers by counts and companies mentioned in synthetic analyst notes.}
\label{fig:ticker_distribution_in_notes}
\end{figure}

The dataset reflects real market structure through industry sector distribution, with Technology sector representing the largest share (31.95\% of notes) followed by Healthcare (9.59\%) and Financial Services (9.07\%), closely matching typical market sector weightings. The note generation process also incorporates market capitalization data to ensure that coverage intensity correlates with company size, as is typical in real analyst coverage.

Each generated note consists of exactly four sentences to ensure consistent length and structure. The note generation process employs templates enriched with numerical metrics and market-specific terminology, with approximately 27.5\% of notes using concept synonyms to increase linguistic variety. Notes can contain up to 4 tickers and 4 concepts each, with special attention paid to maintaining semantic consistency when multiple tickers or concepts appear in the same note, ensuring that the relationships between different entities and concepts are meaningful within the context of each note.

The ticker distribution in our dataset is shown in Fig.~\ref{fig:ticker_distribution_in_notes}.
The distribution of tickers across notes follows patterns typical in analyst coverage, with a higher concentration of single and two-ticker notes, while still maintaining meaningful coverage of more complex multi-ticker scenarios.


Numerical experiments were performed using two datasets of synthetically generated notes. The first dataset has 5000 notes referencing 50 largest stocks in the S\&P500 index, while the second dataset has 10,000 notes referencing about 120 largest stocks. While the results shown below reference the second dataset, we document the performance for both cases.

\subsection{Training CAISSON}

The training of CAISSON was performed on both 5,000 and 10,000 note datasets to assess scalability. Training times were remarkably efficient, taking approximately 60 seconds for the 5,000 note dataset and 120 seconds for the 10,000 note dataset on a standard MacBook Pro laptop, demonstrating near-linear scaling with dataset size.

The training process involves simultaneous optimization of both SOMs, with each maintaining its own learning rate schedule. We used initial learning rates of 0.05 for both SOM1 and SOM2. The batch size was set to 32 documents, providing a good balance between training stability and computational efficiency.

\begin{figure}[h]
\centering
\includegraphics[width=\textwidth]{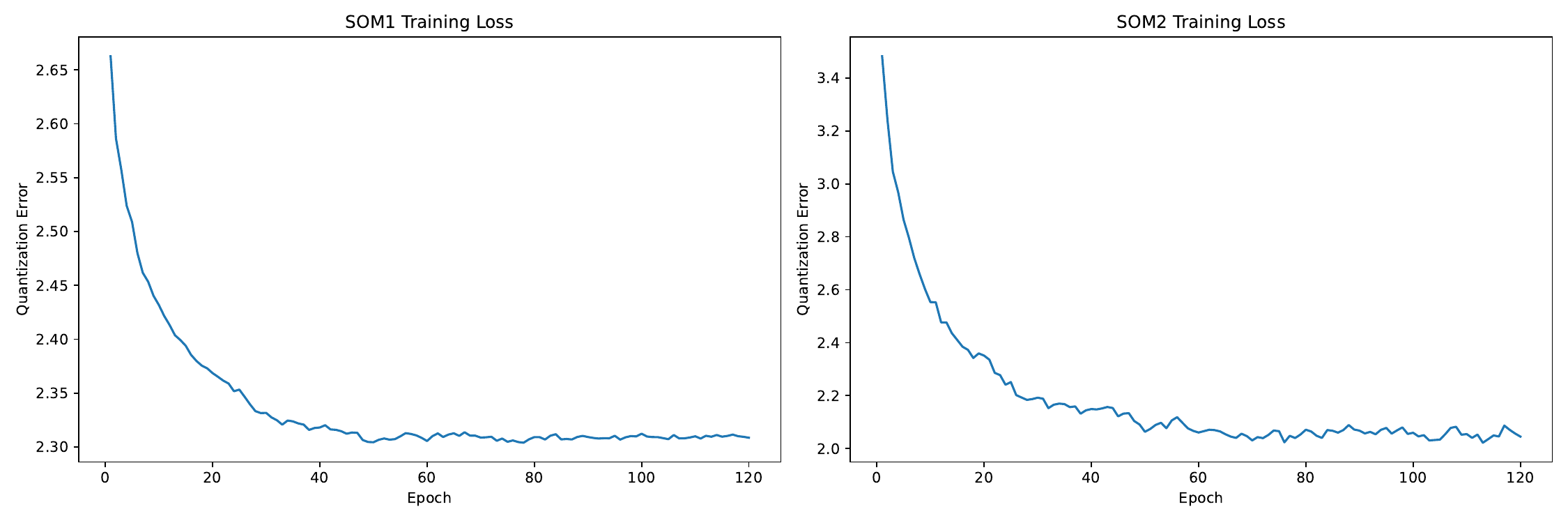}
\caption{Training loss evolution for both SOMs over 150 epochs. The left panel shows the quantization error for SOM1 (text and metadata embeddings), while the right panel shows the same for SOM2 (metadata and concept embeddings). Note the consistent convergence behavior in both cases.}
\label{fig:training_losses}
\end{figure}

The training progress can be monitored through the quantization error of each SOM, as shown in Figure~\ref{fig:training_losses}. Both SOMs demonstrate stable convergence behavior, with the majority of optimization occurring in the first 50 epochs. SOM1, processing the more complex text-metadata embeddings, typically shows a higher initial quantization error but achieves comparable final performance to SOM2.

\begin{figure}[h]
\centering
\includegraphics[width=\textwidth]{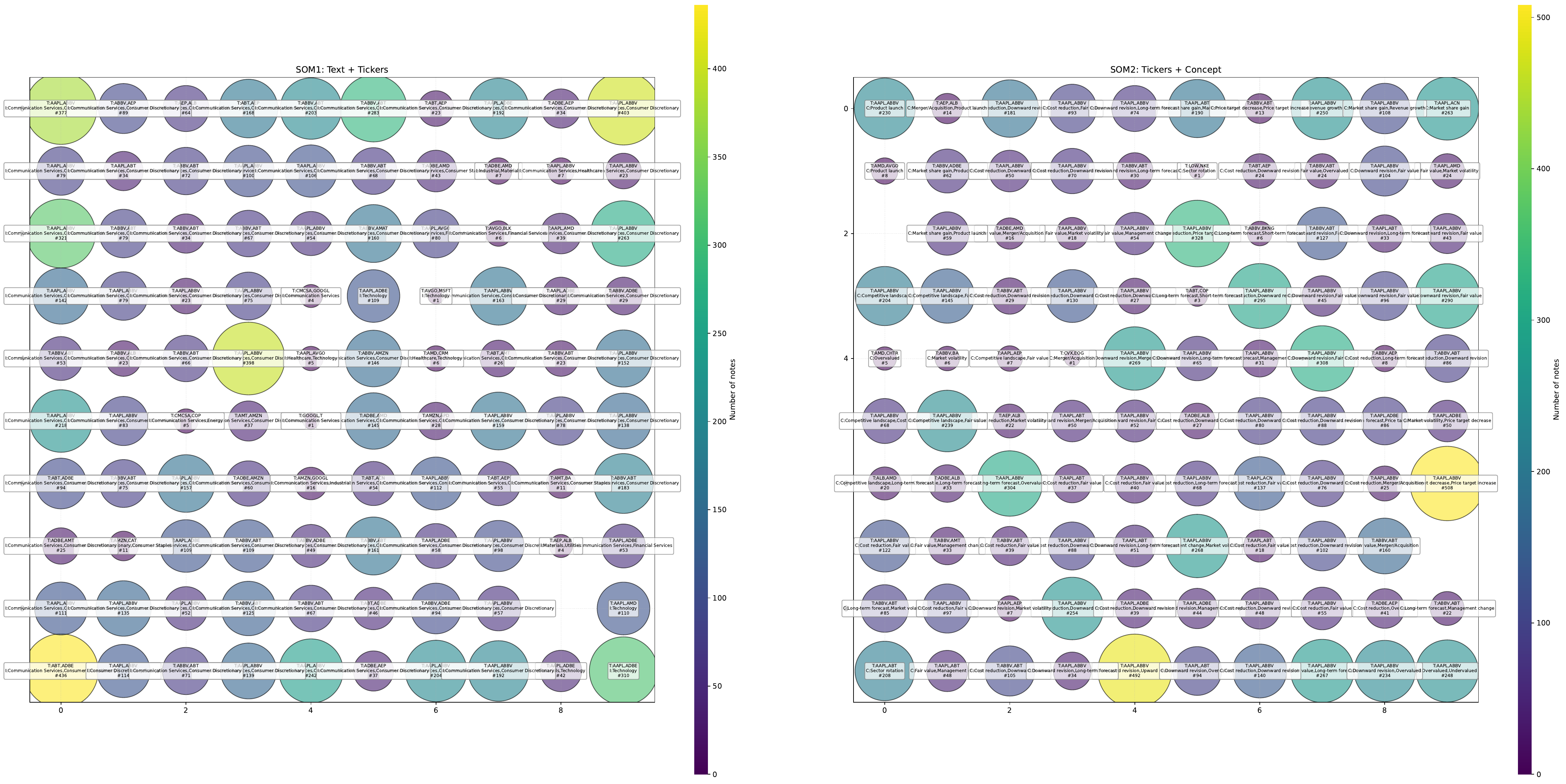}
\caption{Visualization of trained SOMs. Left: SOM1 organization showing clusters of documents based on text and ticker relationships. Right: SOM2 organization displaying concept-based clustering. Node colors indicate cluster density, with darker blues representing higher document concentrations. Labels show representative tickers (T:) and either industries (I:) or concepts (C:) for each cluster.}
\label{fig:som_visualization}
\end{figure}

The effectiveness of the trained model can be visualized through the topological organization of the document space, as illustrated in Figure~\ref{fig:som_visualization}.

The visualization reveals several interesting patterns in the learned document organization:
\begin{itemize}
    \item SOM1 shows clear industry-based clustering, with related companies typically appearing in neighboring nodes
    \item SOM2 exhibits concept-based organization, where similar market events and themes are grouped together
    \item Both SOMs maintain effective separation between different types of market commentary while preserving local similarity relationships
    \item The node activation patterns (indicated by color intensity) show a balanced distribution of documents across the map, suggesting effective use of the topological space
\end{itemize}

The dual-SOM architecture proves particularly effective at capturing both semantic and conceptual relationships, with each SOM specializing in its respective aspect of the document space. This specialization is evident in the distinct clustering patterns observed in Figure~\ref{fig:som_visualization}, where SOM1 shows stronger ticker-based organization while SOM2 reveals clear concept-based groupings.

\section{NoteRetriever: Fast Document Search with CAISSON}
\label{sect_NoteRetriever}

\subsection{Query Processing and Embedding}

The NoteRetriever is a sub-module in CAISSON that implements a sophisticated document search mechanism that leverages both SOMs of the CAISSON model. When a query is received, it undergoes several preprocessing steps to enable effective matching. First, the query text is analyzed to extract relevant tickers using regular expression pattern matching, ensuring that only valid tickers from our universe are considered. Simultaneously, the query is processed to identify relevant concepts, using both direct concept matching and synonym-based detection through a predefined dictionary of concept synonyms.
A more detailed description is given below.

\subsection{Dual-Path Search Strategy}

CAISSON executes parallel searches using both SOMs to leverage their complementary representations. Each query triggers:
\begin{itemize}
\item SOM1 (Text-Metadata Path): Identifies BMU using text-metadata embedding, performs configurable neighborhood search, collects candidate documents from matched nodes.
\item SOM2 (Concept-Metadata Path): Identifies BMU using concept-metadata embedding, executes parallel neighborhood search, generates second candidate set.
\end{itemize}

The dual-path approach enables robust handling of both semantic and conceptual relationships while maintaining computational efficiency through parallel processing.

\subsection{Score Computation and Ranking}

Query processing implements a multi-factor scoring system combining ticker matches, concept similarity, and semantic relevance. For a query $q$ and document $d$, the final score is computed as:
\begin{equation}
Score(q,d) = 0.6 \times TickerScore + 0.2 \times ConceptScore + 0.2 \times SemanticScore
\end{equation}
where:
\begin{align}
TickerScore &= \frac{|T_q \cap T_d|}{\max(|T_q|, |T_d|)}, \
ConceptScore &= sim(C_q, C_d), \
SemanticScore &= \cos(E_q, E_d)
\end{align}
Here $T_q, T_d$ are ticker sets, $C_q, C_d$ are concept sets, and $E_q, E_d$ are document embeddings. This weighting scheme prioritizes exact entity matches while accounting for conceptual and semantic relevance, crucial for financial document retrieval.
The system maintains caches for frequent ticker combinations and concept similarities to optimize retrieval speed, achieving sub-second response times even for complex multi-entity queries.

\subsection{Optimization Techniques}
Three key optimization strategies maintain CAISSON's sub-second response times during document retrieval. First, a comprehensive caching system pre-computes and stores frequently needed information, including concept similarities, multi-entity embeddings, and industry relationships. This significantly reduces computational overhead during query processing by eliminating redundant calculations.

Second, an efficient candidate selection process employs progressive filtering, where exact matches are attempted before falling back to partial matches. The system uses neighborhood-based search strategies around Best Matching Units (BMUs) in both SOMs, configurable based on query complexity. This approach substantially reduces the search space while maintaining retrieval accuracy.

Finally, the implementation leverages optimized similarity computations and memory management techniques. Document vectors are pre-normalized for faster similarity calculations, and all operations are vectorized when possible. The system employs sparse storage strategies and lazy computation patterns, only calculating and storing essential information. These optimizations enable efficient processing of both single-hop and multi-hop queries across datasets of thousands of documents, with typical response times under 200ms for queries involving one to four entities.

\subsection{Example Queries and Retrieved Results}

To illustrate CAISSON's retrieval capabilities, let's examine a specific multi-entity query seeking information about market share developments for two major technology companies:

\begin{mdframed}[frametitle={Query Example}]
Query: "What are the latest developments affecting market share gain for GOOGL and AAPL?"
\end{mdframed}

This query is particularly interesting as it combines multiple elements:
\begin{itemize}
    \item Multiple entities: GOOGL (Alphabet) and AAPL (Apple)
    \item Specific concept: "Market share gain"
    \item Temporal aspect: "latest developments"
\end{itemize}

The system processes this query through both SOMs, producing the following top results:

\begin{mdframed}[frametitle={Retrieved Notes (SOM1)}]
1. "GOOGL has gained 18.1\% market share over the past year. AAPL has gained 5.7\% market share over the past year. GOOGL increased market presence, indicating significant implications for its market position."

2. "AAPL has gained 15.4\% market share over the past year. GOOGL has gained 15.4\% market share over the past year. Recent data shows GOOGL increased market presence, reflecting broader market dynamics."
\end{mdframed}

\begin{mdframed}[frametitle={Retrieved Notes (SOM2)}]
1. "AMZN has gained 9.5\% market share over the past year. ETSY has gained 9.2\% market share over the past year. AMZN competitive position improvement, indicating significant implications for its market position."

2. "COST has gained 0.2\% market share over the past year. HD has gained 3.7\% market share over the past year. Recent data shows HD expanded customer base, reflecting broader market dynamics."
\end{mdframed}

The retrieval results demonstrate several key aspects of CAISSON's capabilities:

1. SOM1 Effectiveness:
  \begin{itemize}
      \item Successfully identifies notes containing both queried entities (GOOGL and AAPL)
      \item Maintains semantic relevance to market share discussions
      \item Preserves temporal context ("over the past year")
  \end{itemize}

2. SOM2 Behavior:
  \begin{itemize}
      \item Focuses on conceptually similar notes about market share gains
      \item Includes related companies from similar sectors (e.g., AMZN, ETSY in e-commerce)
      \item Maintains the same semantic structure even with different entities
  \end{itemize}

\section{SynFAQA: Evaluation Framework for CAISSON}
\label{sect_SynFAQA}

\subsection{Generation of Question-Answer Pairs}

Building upon our synthetic analyst notes dataset described in Section~\ref{sect_CAISSON_test}, we developed SynFAQA (Synthetic Financial Analyst Questions and Answers generator): a comprehensive evaluation framework that tests both single-hop and multi-hop question retrieval capabilities of CAISSON. Using synthetic data as the foundation for our evaluation framework provides a crucial advantage: we have complete knowledge of the underlying information and relationships in the documents, enabling creation of questions with verifiable ground truth answers. This contrasts with evaluations based on real financial documents, where establishing ground truth often requires extensive manual annotation and may still contain ambiguities.

Drawing inspiration from HotPotQA~\cite{Yang2018hotpotqa}, our framework extends their methodology to the financial domain, incorporating domain-specific concepts and entity relationships while maintaining the controlled nature of our synthetic data. The resulting framework may be of broader interest for benchmarking LLMs on financial text analysis tasks, as it provides a systematic way to evaluate both retrieval accuracy and reasoning capabilities with known ground truth.

The question generation process employs a graph-based approach where notes are connected through bridge elements - either shared tickers or common concepts. This enables the generation of three distinct types of questions:

\begin{itemize}
    \item Bridge questions: Require finding connections between notes through common elements
    \item Comparison questions: Focus on comparing similar metrics or developments across different entities
    \item Yes/No questions: Ask about the similarity or difference in trends between entities
\end{itemize}

%
%

\subsection{Question Types and Distribution}

To ensure comprehensive evaluation of CAISSON's capabilities, we generated a balanced dataset of 20,000 questions evenly split between single-hop and multi-hop queries (Table~\ref{tab:question_dist}). Single-hop questions test basic retrieval capabilities with queries that can be answered using information from a single document. Multi-hop questions evaluate more sophisticated retrieval abilities by requiring information synthesis across multiple documents.The multi-hop questions are further categorized into three types based on their reasoning requirements, see Table~\ref{tab:multihop_dist}).

\begin{table}[h]
\centering
\begin{tabular}{|l|r|r|}
\hline
Question Type & Count & Percentage \\
\hline
Single-hop & 10,000 & 50\% \\
Multi-hop & 10,000 & 50\% \\
\hline
\end{tabular}
\caption{Distribution of question types in the evaluation dataset}
\label{tab:question_dist}
\end{table}

\begin{table}[h]
\centering
\begin{tabular}{|l|r|r|}
\hline
Multi-hop Type & Count & Percentage \\
\hline
Bridge & 6,005 & 60.1\% \\
Yes/No & 2,963 & 29.6\% \\
Comparison & 1,032 & 10.3\% \\
\hline
\end{tabular}
\caption{Distribution of multi-hop question types}
\label{tab:multihop_dist}
\end{table}

\subsection{Bridge Element Analysis}

Bridge elements are crucial components of multi-hop questions, providing the logical connections between related documents. Our analysis focused on validating these connections to ensure the questions genuinely require multi-document synthesis rather than being answerable through single-document retrieval. Throughout our question set, we identified and validated 1,730 distinct concept-based bridges, with each bridge element verified to exist in its respective connected documents. This validation process ensures that our multi-hop questions provide a meaningful test of CAISSON's ability to perform complex information synthesis across multiple documents, aligning with its dual-view architecture that explicitly handles both semantic and conceptual relationships.

%
%
%
%
%
%

\subsection{Examples and Question Difficulty Analysis}
Our framework generates questions of varying complexity. Here are representative examples of different question types:

\begin{itemize}
   \item Single-hop question:\\
   "What's the latest information on TSLA regarding upward revision?"\\
   This requires finding relevant information about TSLA's earnings revisions from a single note.

   \item Bridge question:\\
   "What different approaches do BA and BKNG take regarding enhanced operational margins?"\\
   This question requires bridging information about profit margin expansion across two different notes about BA and BKNG.

   \item Comparison question:\\
   "Between PXD and COP, which company had more favorable fair value?"\\
   This involves comparing valuation metrics between two energy sector companies.

   \item Yes/No question:\\
   "Did EQR and WELL experience similar trends in product launch?"\\
   This requires analyzing and comparing product launch information across two real estate companies.
   
   \item Multi-ticker questions:\\
    Some of questions have up to 4 tickers mentioned in them. Examples of such 
   questions: \\
   "Did SHW, FCX, and DOW and NEM experience similar trends in price target increase?" \\
   "What insights emerge when comparing industry rivalry situation between LLY and PG and KO and MRK?" \\
   Such queries might be quite challenging due to their high ticker complexity.
   
\end{itemize}



\subsection{Question Complexity Analysis}

We analyzed the complexity of generated questions across multiple dimensions, including syntactic features (word and character counts) and semantic complexity (number of entities and concepts). Results are shown in Table~\ref{tab:complexity}.

\begin{figure}[h]
\centering
\includegraphics[height=9cm, width=12cm]{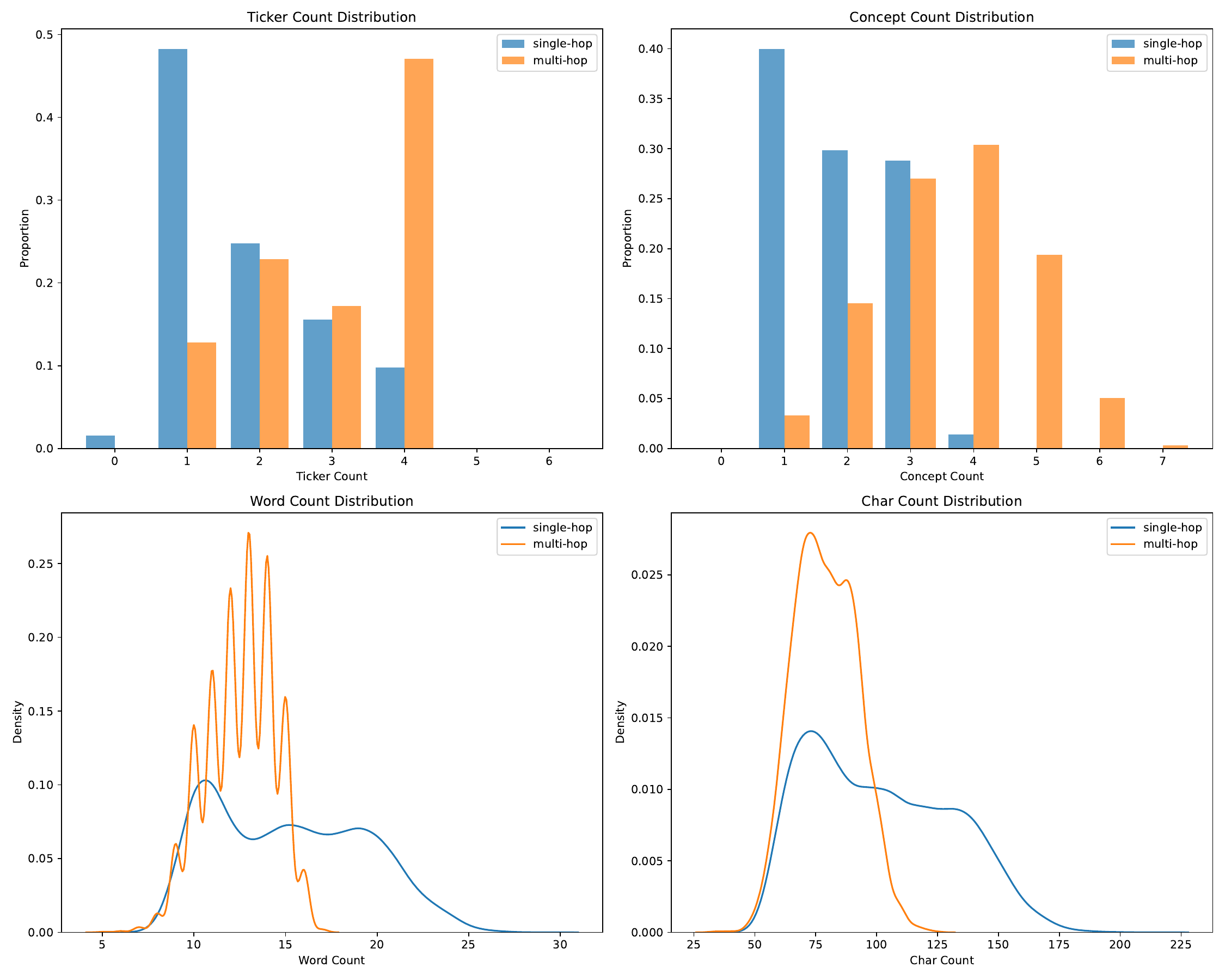}
\caption{Distributions of tickers, concepts, words and character counts  for single-hop and multi-hop questions.}
\label{fig:complexity_metrics}
\end{figure}

\begin{figure}[h]
\centering
\includegraphics[height=6cm, width=8cm]{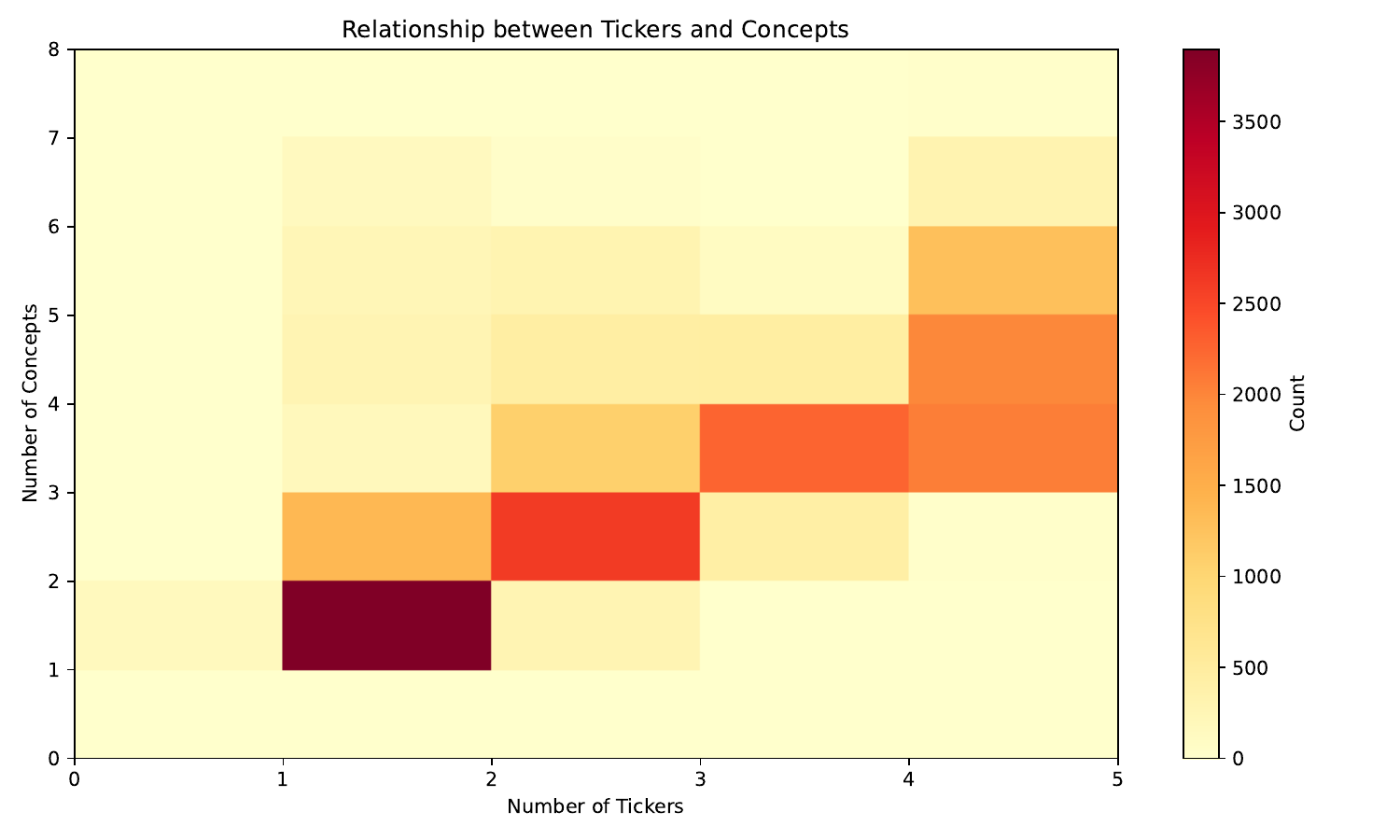}
\caption{Query complexity in terms of tickers-concepts counts.}
\label{fig:ticker_concept}
\end{figure}

\begin{table}[h]
    \caption{Question complexity metrics}
    \label{tab:complexity}
    \begin{center}
    \begin{tabular}{|c|c|c|c|c|}
    \hline
    Type & Metric & Mean & Std & Range \\
    \hline
single-hop & word\_count & 15.22 & 4.14 & 8-29 \\
single-hop & char\_count & 101.65 & 29.21 & 51-214 \\
single-hop & ticker\_count & 1.84 & 1.03 & 0-4 \\
single-hop & concept\_count & 1.92 & 0.86 & 1-4 \\
multi-hop & word\_count & 12.53 & 1.86 & 5-17 \\
multi-hop & char\_count & 79.09 & 12.89 & 32-126 \\
multi-hop & ticker\_count & 2.99 & 1.10 & 0-4 \\
multi-hop & concept\_count & 3.64 & 1.20 & 1-7 \\
    \hline
    \end{tabular}
    \end{center}
    \end{table}
The analysis reveals distinct patterns between single-hop and multi-hop questions:
\begin{itemize}
   \item Single-hop questions tend to be longer (mean 15.22 words) but more focused, with fewer entities (mean 1.84 tickers) and concepts (mean 1.92)
   \item Multi-hop questions are more concise (mean 12.53 words) but denser in information content, containing more entities (mean 2.99 tickers) and concepts (mean 3.64)
\end{itemize}
Fig.~\ref{fig:complexity_metrics} provides detailed distributions of structural features, while Fig.~\ref{fig:ticker_concept} illustrates the relationship between ticker and concept complexity. 
    
Questions were further classified into difficulty levels based on their structural complexity:
\begin{itemize}
   \item Easy: Single-entity queries or questions with one concept
   \item Medium: Two-entity queries or questions involving two concepts
   \item Hard: Questions with three or more entities, or more than two concepts
\end{itemize}

As shown in Table~\ref{tab:difficulty}, this classification reveals a stark contrast between single-hop and multi-hop questions. Single-hop questions show a relatively balanced distribution across difficulty levels (40.0\% easy, 29.8\% medium, 30.2\% hard). In contrast, multi-hop questions are predominantly hard (86.7\%), with very few easy cases (0.5\%), reflecting their inherently more complex nature requiring information synthesis across multiple documents.
This enables systematic evaluation of retrieval systems across different complexity levels and helps identify potential areas for optimization.

\begin{table}[h]
    \caption{Question difficulty distribution}
    \label{tab:difficulty}
    \begin{center}
    \begin{tabular}{|l|c|c|c|}
    \hline
    Question Type & Easy & Medium & Hard \\
    \hline
single-hop & 40.0\% & 29.8\% & 30.2\% \\
multi-hop & 0.5\% & 12.8\% & 86.7\% \\
    \hline
    \end{tabular}
    \end{center}
    \end{table}


\section{Experimental Evaluation and Benchmark Comparisons with SynFAQA}
\label{sect_CAISSON_evaluation}

Our evaluation of CAISSON focuses on three key aspects: comparison against baseline RAG implementations, analysis of performance across different question types, and assessment of scaling behavior with query complexity. We begin by establishing strong and weak baselines to isolate the specific contributions of CAISSON's architectural innovations, then proceed to detailed performance analysis across different dimensions of retrieval tasks.

\subsection{Baseline RAG Models}

To rigorously evaluate CAISSON's performance, we implement two strategically chosen baseline models that represent different levels of architectural sophistication:

\begin{itemize}
    \item \textbf{TextRAG (Weak Baseline):} This implementation represents the traditional RAG approach, using only text embeddings for retrieval. It serves as a basic benchmark that helps quantify improvements over standard industry practice. TextRAG:
    \begin{itemize}
        \item Uses pure text embeddings without entity awareness
        \item Implements standard vector similarity search
        \item Represents the common baseline in RAG literature
    \end{itemize}
    
    \item \textbf{TextEntityRAG (Strong Baseline):} This enhanced implementation replicates key components of CAISSON's entity-handling capabilities, making it a particularly challenging benchmark. Specifically, TextEntityRAG:
    \begin{itemize}
        \item Uses concatenated text and entity embeddings, matching CAISSON's first SOM path
        \item Implements sophisticated metadata-aware filtering based on ticker information
        \item Employs the same query processing logic for entity handling as CAISSON
        \item Effectively replicates CAISSON's entity-awareness mechanisms
    \end{itemize}
\end{itemize}

TextEntityRAG serves as an especially important benchmark because it isolates the specific contribution of CAISSON's multi-view architecture. By implementing sophisticated entity handling but maintaining a single-view approach, TextEntityRAG helps us quantify the additional value provided by CAISSON's concept-handling mechanism and dual-SOM organization. Any performance improvements over TextEntityRAG can be directly attributed to these architectural innovations rather than simple entity awareness.

All models use the same underlying MiniLM-L6-v2 text embedding model to ensure fair comparison. Evaluation is performed on a set of 1,000 randomly selected questions (500 single-hop and 500 multi-hop). The evaluation framework assesses several key aspects of the model's performance:

\begin{itemize}
   \item Retrieval accuracy at different ranks (top-1, top-5, and top-10)
   \item Mean Reciprocal Rank (MRR)
   \item Ticker and concept coverage in retrieved documents
   \item Processing time
\end{itemize}

\subsection{Overall Performance}

\begin{figure}[t]
\centering
\includegraphics[width=\linewidth]{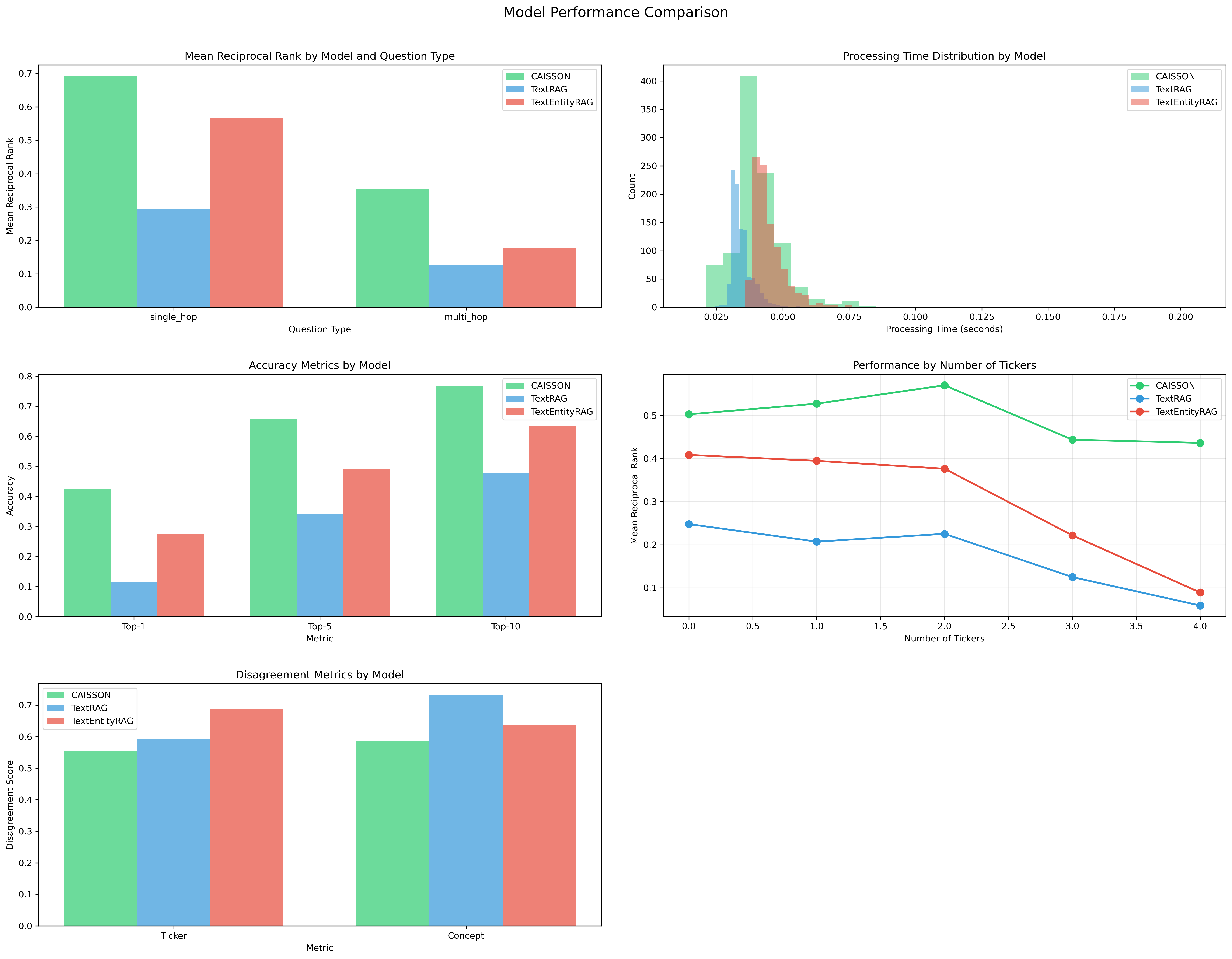}
\caption{Detailed performance comparison across models showing MRR, accuracy metrics, and disagreement scores.}
\label{fig:model_comparison}
\end{figure}

Tables~\ref{tab:model_comparison} and Fig.~\ref{fig:model_comparison} present the comprehensive performance metrics. CAISSON demonstrates superior retrieval performance even against the strong TextEntityRAG benchmark, achieving a Mean Reciprocal Rank (MRR) of 0.5231 compared to 0.3720 for TextEntityRAG (40.6\% improvement) and 0.2106 for the basic TextRAG (148.4\% improvement). The substantial improvement over TextEntityRAG is particularly significant as it demonstrates that CAISSON's multi-view clustering approach provides benefits beyond what can be achieved through sophisticated entity handling alone.

\begin{table}[t]
\caption{Overall Model Performance}
\label{tab:overall_performance}
\label{tab:model_comparison}
\begin{center}
\begin{tabular}{|c| c| c| c| c| c| c| c|}
\hline
Model & MRR & Top-1 & Top-5 & Top-10 & Time (ms) & Ticker Dis. & Concept Dis. \\
\hline
CAISSON & 0.5231 & 0.4240 & 0.6580 & 0.7680 & 40.4 & 0.5537 & 0.5852 \\
TextRAG & 0.2106 & 0.1140 & 0.3430 & 0.4780 & 34.5 & 0.5929 & 0.7315 \\
TextEntityRAG & 0.3720 & 0.2740 & 0.4920 & 0.6350 & 44.9 & 0.6879 & 0.6361 \\
\hline
\end{tabular}
\end{center}
\end{table}
Notably, CAISSON achieves lower disagreement scores for both ticker (0.5537 vs 0.5929/0.6879) and concept matching (0.5852 vs 0.7315/0.6361), indicating more precise entity and concept coverage in retrieved documents. This improved accuracy comes with only modest computational overhead (40.4 ms vs 34.5/44.9 ms), demonstrating the efficiency of the dual-SOM architecture.

\subsection{Question Type Analysis}

Table~\ref{tab:question_type_performance} compares CAISSON's performance against baseline models across different question types. CAISSON shows particularly strong performance on single-hop queries, with an MRR of 0.6911 compared to 0.2947 for TextRAG and 0.5655 for TextEntityRAG. For multi-hop queries, while the performance somewhat deteriorates, CAISSON still maintains superior retrieval quality with an MRR of 0.3551 versus 0.1265/0.1786 for the baseline models.

A more detailed analysis of multi-hop question types (Table~\ref{tab:multi_hop}) reveals that bridge questions achieve the strongest performance (MRR 0.4209, Top-5 accuracy 0.6808), followed by Yes/No questions (MRR 0.3504) and comparison questions (MRR 0.2468). This pattern suggests that CAISSON's dual-path architecture is particularly effective at handling queries requiring connection identification between documents.

%

\begin{table}[t]
\caption{Comparison of CAISSON against baseline models for different question types}
\label{tab:question_type_performance}
\begin{center}
\begin{tabular}{|c| c| c| c| c| c|}
\hline
Model & Question Type & MRR & Top-5 & Ticker Dis. & Concept Dis. \\
\hline
CAISSON & Single Hop & 0.6911 & 0.7620 & 0.5796 & 0.5357 \\
CAISSON & Multi Hop & 0.3551 & 0.5540 & 0.5277 & 0.6346 \\
TextRAG & Single Hop & 0.2947 & 0.4400 & 0.6048 & 0.7175 \\
TextRAG & Multi Hop & 0.1265 & 0.2460 & 0.5810 & 0.7455 \\
TextEntityRAG & Single Hop & 0.5655 & 0.6900 & 0.7274 & 0.5995 \\
TextEntityRAG & Multi Hop & 0.1786 & 0.2940 & 0.6484 & 0.6727 \\
\hline
\end{tabular}
\end{center}
\end{table}

\begin{table}[htbp]
\centering
\begin{tabular}{|c|c|c|c|c|c|c|}
\hline
Subtype & MRR & Top-1 & Top-5 & Top-10 & Ticker Dis. & Concept Dis. \\
\hline
Bridge & 0.4209 & 0.2508 & 0.6808 & 0.8013 & 0.5629 & 0.6089 \\\hline
Comparison & 0.2468 & 0.0909 & 0.4394 & 0.7273 & 0.4907 & 0.6261 \\\hline
Yes\_No & 0.3504 & 0.2047 & 0.5197 & 0.6850 & 0.4892 & 0.6923 \\\hline
\end{tabular}
\caption{Detailed performance analysis of CAISSON across different multi-hop question types}
\label{tab:multi_hop}
\end{table}
Key findings from this analysis include:
\begin{itemize}
   \item While single-hop performance (MRR 0.6911) significantly exceeds multi-hop performance (MRR 0.3551), both show substantial improvements over baseline models
   \item Performance remains robust even for challenging queries, with Top-5 accuracy exceeding 0.43 across all question types
   \item Concept and ticker disagreement scores remain stable across question types, indicating consistent entity tracking
   \item Processing time remains efficient across all question types, enabling interactive use
\end{itemize}

\subsection{Ticker Complexity Analysis}

Having established CAISSON's overall performance advantages, we now examine how the system handles queries of varying complexity. This analysis is particularly important given the wide range of query types encountered in real-world financial applications.
Table~\ref{tab:ticker_complexity} (see also Fig.~\ref{fig:model_comparison}) reveals an interesting relationship between performance and query complexity as measured by the number of tickers.
\begin{table}[t]
\caption{Performance by Number of Tickers}
\label{tab:ticker_complexity}
\begin{center}
\begin{tabular}{|c| c| c| c| c|}
\hline
Model & Tickers & MRR & Ticker Dis. & Concept Dis. \\
\hline
CAISSON & 0 & 0.5029 & 0.5286 & 0.4894 \\
CAISSON & 1 & 0.5276 & 0.5425 & 0.5663 \\
CAISSON & 2 & 0.5703 & 0.5989 & 0.6721 \\
CAISSON & 3 & 0.4439 & 0.5670 & 0.6916 \\
CAISSON & 4 & 0.4366 & 0.5042 & 0.6278 \\
TextRAG & 0 & 0.2479 & 0.5727 & 0.6570 \\
TextRAG & 1 & 0.2071 & 0.5793 & 0.7413 \\
TextRAG & 2 & 0.2253 & 0.6442 & 0.7695 \\
TextRAG & 3 & 0.1248 & 0.5724 & 0.7822 \\
TextRAG & 4 & 0.0587 & 0.5868 & 0.7353 \\
TextEntityRAG & 0 & 0.4086 & 0.7009 & 0.5398 \\
TextEntityRAG & 1 & 0.3949 & 0.6778 & 0.6472 \\
TextEntityRAG & 2 & 0.3765 & 0.7223 & 0.6879 \\
TextEntityRAG & 3 & 0.2217 & 0.6211 & 0.6914 \\
TextEntityRAG & 4 & 0.0889 & 0.6223 & 0.6743 \\
\hline
\end{tabular}
\end{center}
\end{table}
Several key patterns emerge:
\begin{itemize}
    \item \textbf{Consistent Strong Performance:} CAISSON maintains robust performance across different ticker complexities, with MRR scores ranging from 0.5029 (no tickers) to 0.5703 (two tickers). This significantly outperforms both TextRAG (MRR range 0.0587-0.2479) and TextEntityRAG (MRR range 0.0889-0.4086).
    
    \item \textbf{Peak Performance at Moderate Complexity:} Interestingly, CAISSON achieves its highest performance with two-ticker queries (MRR 0.5703), showing that the dual-SOM architecture effectively handles moderate complexity queries. This represents a 153\% improvement over TextRAG (MRR 0.2253) and 51.5\% over TextEntityRAG (MRR 0.3765) for the same query type.
    
    \item \textbf{Graceful Performance Scaling:} While performance does decrease with higher ticker counts, CAISSON maintains strong performance even for three and four-ticker queries (MRR 0.4439 and 0.4366 respectively), substantially outperforming both baseline models which show more dramatic performance degradation (TextRAG: 0.1248 and 0.0587; TextEntityRAG: 0.2217 and 0.0889).
    
    \item \textbf{Robust Entity and Concept Handling:} CAISSON shows more consistent disagreement scores across all complexity levels (ticker disagreement range 0.5042-0.5989) compared to the more variable performance of TextRAG (0.5724-0.6442) and TextEntityRAG (0.6211-0.7223), indicating more stable entity tracking regardless of query complexity.
\end{itemize}

\subsection{Implications for Real-World Applications}

The performance characteristics of CAISSON reveal several important implications for practical applications:

\begin{itemize}
    \item \textbf{Robust Query Handling:} CAISSON demonstrates exceptional performance across a wide range of query complexities, with strong results for queries involving up to four tickers. This makes it suitable for real-world financial applications, where such queries represent the majority of analyst information needs.
    
    \item \textbf{Enhanced Multi-Entity Analysis:} The system's peak performance with two-ticker queries (MRR 0.5703) suggests particular strength in comparative analysis scenarios, a crucial capability for financial applications where understanding relationships between multiple entities is essential.
    
    \item \textbf{Scalable Performance:} Unlike traditional RAG approaches that show sharp performance degradation with increased complexity, CAISSON maintains strong performance even for four-ticker queries (MRR 0.4366), demonstrating robust scalability for complex analytical tasks.
    
    \item \textbf{Balanced Trade-offs:} While CAISSON's processing time is modestly higher than baseline approaches, the substantial improvements in retrieval accuracy (often exceeding 100\% for complex queries) justify this trade-off for applications where precision is crucial.
    
    \item \textbf{Enterprise Readiness:} The combination of strong multi-entity performance, consistent entity tracking (demonstrated by stable disagreement scores), and sub-second response times makes CAISSON particularly well-suited for enterprise-grade financial analysis platforms.
\end{itemize}

\section{Summary and Future Directions}
\label{sect_Summary}

\subsection{Summary of Key Results}

This paper introduced CAISSON as a novel approach to RAG systems that combines multi-view clustering with document retrieval. By leveraging dual Self-Organizing Maps, CAISSON creates complementary organizational views of the document space - one based on semantic-entity relationships and another on concept-entity patterns. This multi-view clustering approach enables more nuanced and accurate document retrieval while maintaining efficient query processing. Our comprehensive evaluation demonstrates several key findings:

\begin{itemize}
    \item \textbf{Enhanced Retrieval Performance:} CAISSON achieves substantial improvements over baseline RAG models:
    \begin{itemize}
        \item 148.4\% improvement in Mean Reciprocal Rank (0.5231 vs 0.2106) compared to TextRAG
        \item 40.6\% improvement in MRR (0.5231 vs 0.3720) over the strong TextEntityRAG benchmark
        \item Exceptional Top-1 accuracy (0.4240 vs 0.1140/0.2740)
        \item Strong Top-5 accuracy (0.6580 vs 0.3430/0.4920)
    \end{itemize}
    
    \item \textbf{Query Type and Complexity Handling:}
    \begin{itemize}
        \item Outstanding single-hop performance (MRR 0.6911)
        \item Strong multi-hop capabilities (MRR 0.3551)
        \item Peak performance for two-ticker queries (MRR 0.5703)
        \item Robust performance maintained through four-ticker queries (MRR 0.4366)
    \end{itemize}
    
    \item \textbf{Multi-View Clustering Benefits:} 
    \begin{itemize}
        \item Effective organization of document space through complementary views
        \item Superior entity tracking across query complexities
        \item Enhanced concept-aware retrieval through dedicated SOM path
        \item Balanced performance across different query types
    \end{itemize}
\end{itemize}

A particularly significant finding is that CAISSON's multi-view clustering approach substantially outperforms both basic RAG implementations and enhanced entity-aware systems. The performance advantages are especially pronounced for multi-entity queries, where the dual-path architecture enables more sophisticated handling of complex relationships. These results demonstrate the value of combining classical clustering techniques with modern embedding approaches for document retrieval.

\subsection{Connections to Recent Developments}

Our approach to evaluation and the observed performance patterns align with several recent developments in AI research. The SynFAQA methodology shares important conceptual parallels with DeepMind's Michelangelo framework~\cite{DeepMind2024michelangelo}, particularly in using controlled synthetic data for systematic evaluation. While Michelangelo targets general long-context reasoning, our framework applies similar principles to financial domain reasoning, with both approaches enabling precise assessment of model capabilities through:

\begin{itemize}
    \item \textbf{Controlled Task Design:} Our single-hop and multi-hop questions mirror Michelangelo's approach to primitive tasks, allowing systematic evaluation of both simple retrieval and complex reasoning capabilities. This is evidenced by CAISSON's strong differentiated performance across query types, achieving MRR of 0.6911 for single-hop queries while maintaining robust performance (MRR 0.3551) for more complex multi-hop queries.
    
    \item \textbf{Complexity Scaling:} The controlled progression of query complexity provides systematic assessment of model capabilities, similar to Michelangelo's complexity scaling approach. Our results show CAISSON's particular strength in moderate complexity queries (peak MRR 0.5703 for two-ticker queries) while maintaining strong performance (MRR $ > 0.43$) even for complex four-ticker queries.
    
    \item \textbf{Domain-Specific Benchmarking:} Our synthetic data generation maintains financial domain characteristics while preventing training data leakage, enabling reliable comparison of different RAG architectures on realistic tasks. The framework's ability to generate controlled test cases with varying complexity enables systematic evaluation of both retrieval accuracy and scaling behavior.
\end{itemize}

\subsection{Future Research Directions}

Our findings suggest several promising directions for future research, prioritized by their potential impact:

\begin{enumerate}
    \item \textbf{Architecture Extensions:}
    \begin{itemize}
        \item Development of adaptive view-weighting mechanisms for different query types
        \item Extension of the multi-view approach to capture temporal relations in the data
        \item Extension to longer documents where document-level concepts and metadata must inform chunk-level retrieval
    \end{itemize}
    
    \item \textbf{Enhanced Concept Processing:} 
    \begin{itemize}
        \item Zero-shot concept detection using Kimhi et al.'s~\cite{Kimhi2022interpreting} conceptualization approach
        \item Dynamic concept discovery building on Anthropic's contextual embeddings~\cite{Anthropic2024contextual}
        \item Integration with domain-specific ontologies for richer concept representation
    \end{itemize}
    
    \item \textbf{Production and Domain Extensions:}
    \begin{itemize}
        \item Distributed implementations for large-scale document collections
        \item Efficient updating strategies for dynamic content
        \item Adaptation to domains with rich entity relationships (scientific literature, legal documents)
    \end{itemize}
\end{enumerate}

A particularly promising direction would be the development of adaptive multi-view architectures that automatically optimize their organization based on both document characteristics and domain structure. This could enable effective handling of more complex document collections while maintaining CAISSON's efficient retrieval capabilities.

\subsection{Broader Implications}

The success of CAISSON has significant implications for both RAG system design and document retrieval architectures. Our results demonstrate that carefully designed multi-view clustering approaches can substantially enhance retrieval capabilities beyond what is achievable through either simple embedding concatenation or sophisticated single-view organizations. The clear performance advantage over TextEntityRAG (40.6\% improvement in MRR) is particularly noteworthy, as it demonstrates the value of concept-based organizational views beyond entity awareness alone.

Looking beyond our current experimental setting with short analyst notes, we anticipate that CAISSON's dual-view architecture will offer even greater advantages for complex document collections. As document length and complexity increase, the ability to efficiently track and retrieve concept-related information becomes increasingly crucial, as relevant conceptual patterns may be distributed across larger spans of text. The multi-view approach provides a natural framework for handling this increased complexity while maintaining retrieval efficiency.

More broadly, our results suggest a promising direction for next-generation RAG systems through the combination of classical clustering techniques with modern embedding approaches. This hybrid approach could be particularly valuable in domains where documents exhibit multiple types of relationships and where efficient multi-aspect retrieval is crucial for practical applications. As the field continues to evolve, we believe CAISSON's principles of topology-preserving organization and multi-view document representation will prove increasingly relevant for handling the complexity of real-world information retrieval tasks.

\bibliographystyle{plain}
\bibliography{references}

\end{document}